\documentclass[10pt,twocolumn,letterpaper]{article}

\usepackage{cvpr}      

\usepackage{graphicx}
\usepackage{amsmath}
\usepackage{amssymb}

\usepackage{booktabs}
\usepackage{multirow}
\usepackage{bm}
\usepackage{subcaption}
\usepackage{enumitem}
\usepackage{threeparttable}
\usepackage[table,xcdraw]{xcolor}
\usepackage[pagebackref,breaklinks,colorlinks]{hyperref}

\usepackage[capitalize]{cleveref}
\crefname{section}{Sec.}{Secs.}
\Crefname{section}{Section}{Sections}
\Crefname{table}{Table}{Tables}
\crefname{table}{Tab.}{Tabs.}

\usepackage{color,soul}
\usepackage{xcolor}

\newcommand{\yf}[1]{\textcolor{black}{#1}}
\newcommand{\wh}[1]{\textcolor{black}{#1}}

\def\cvprPaperID{2717} 
\def\confName{CVPR}
\def\confYear{2023}

\begin{document}

\title{Raw Image Reconstruction with Learned Compact Metadata}
\author{Yufei Wang$^1$, Yi Yu$^1$, Wenhan Yang$^2$, Lanqing Guo$^1$, Lap-Pui Chau$^3$, Alex C. Kot$^1$, Bihan Wen$^1\thanks{Corresponding author.}$\\
$^1$Nanyang Technological University \quad $^2$Peng Cheng Lab\\
$^3$The Hong Kong Polytechnic University \\
{\tt\small \{yufei001, yuyi0010, lanqing001, eackot, bihan.wen\}@ntu.edu.sg}\\ {\tt\small yangwh@pcl.ac.cn \quad lap-pui.chau@polyu.edu.hk}
}
\maketitle

\begin{abstract}
While raw images exhibit \wh{advantages} over sRGB images \wh{(\textit{e.g.,} linearity and fine-grained quantization level)}, they are not widely used by common users due to the large storage requirements.
%
Very recent works propose to compress raw images by designing the sampling masks in the raw image pixel space, leading to suboptimal image representations and redundant metadata. 
%
%
In this paper, we propose \wh{a novel framework} to learn a compact representation in the latent space serving as the metadata \wh{in} an end-to-end manner.
%
Furthermore, we propose a novel sRGB-guided context model with the improved entropy estimation strategies, which leads to better reconstruction quality, smaller size of metadata, and faster speed.
We illustrate how the proposed raw image compression scheme can adaptively allocate more bits to image regions that are important from a global perspective.
%
%
The experimental results show that the proposed method can achieve superior raw image reconstruction results using a smaller size of the metadata on both uncompressed sRGB images and JPEG images. The code will be released at \url{https://github.com/wyf0912/R2LCM}.
\end{abstract}

\section{Introduction}
As an unprocessed and uncompressed data format \wh{directly obtained from camera sensors}, raw images provide unique advantages for computer vision tasks in practice.
%
\wh{For example, it is easier to model the distribution of real image noise in raw space, which enables generalized deep real denoising networks~\cite{zhang2021rethinking, abdelhamed2019noise};}
%
%
\wh{As pixel values in raw images have a linear relationship with scene radiance, they own benefits to recover shadows and highlights without bringing in the grainy noise usually associated with high ISO, which greatly contributes to the low-light image enhancement~\cite{wei2020physics, huang2022towards, wang2022low}.}
%
{Besides, with richer colors, raw images offer more room for correction and artistic manipulation.}

\begin{figure}[t]
    \centering
     \begin{subfigure}{1\linewidth}
    \includegraphics[width=1\linewidth]{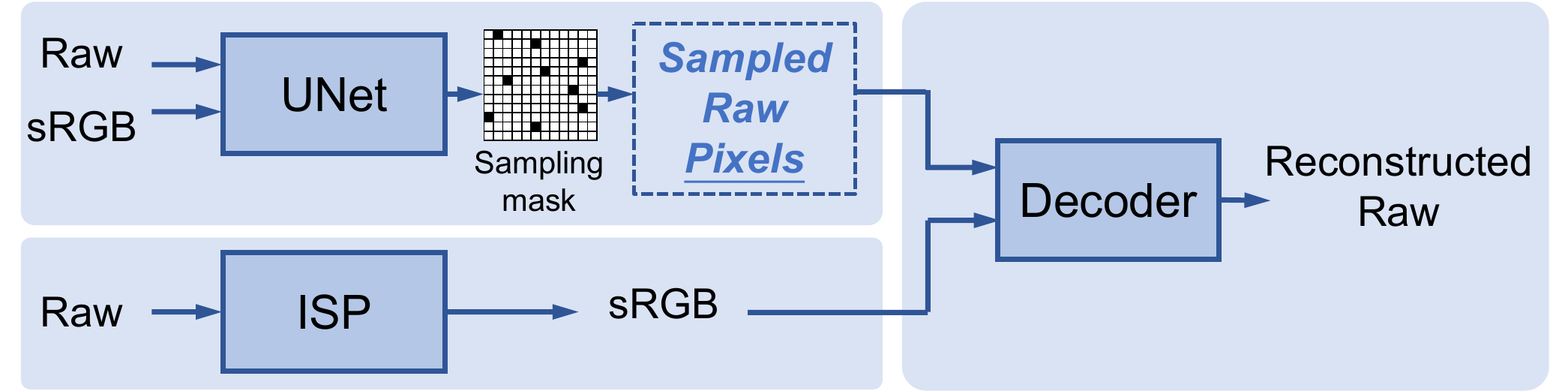}
    \caption{Previous SOTA methods sample in the raw space \cite{punnappurath2021spatially, nam2022learning}.}
     \end{subfigure}
     \begin{subfigure}{1\linewidth}
    \includegraphics[width=1\linewidth]{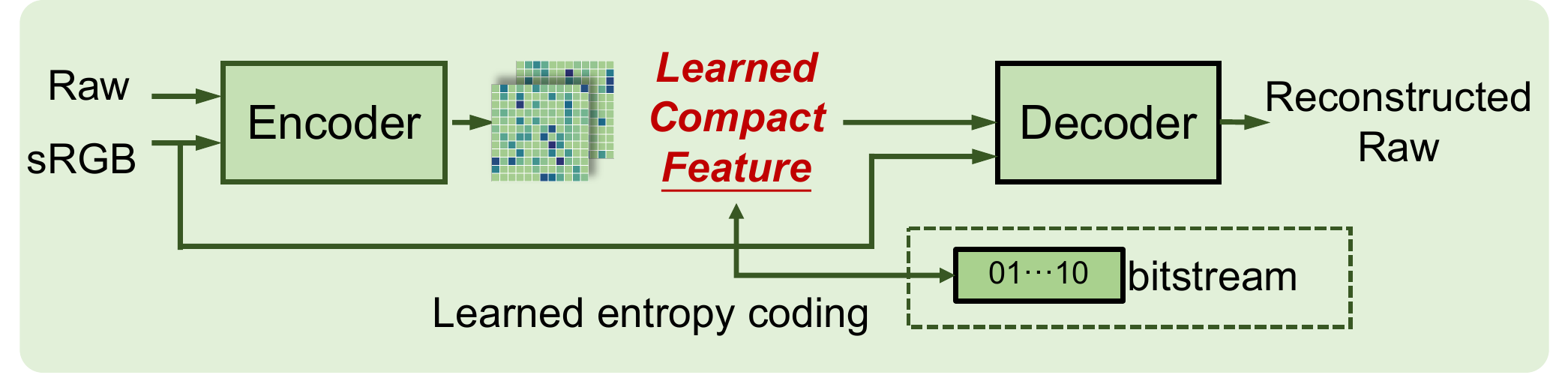}
    \vspace{-0.3cm}
    \caption{The proposed method samples in the latent space.}
     \end{subfigure}
    \vspace{-0.4cm}
    \caption{
    The comparison between the previous SOTA methods (in blue) and our proposed method (in green).
    Different from the previous work where the sampling strategy is hand-crafted or learned by a pre-defined sampling loss, we learn the sampling and reconstruction process in a unified end-to-end manner. 
    In addition, the sampling of previous works is in the raw pixel space, which in fact still includes a large amount of spatial redundancy and precision redundancy.
    Instead, we conduct sampling in the feature space, and more compact metadata is obtained for pixels in the feature space via the adaptive allocation. The saved metadata is annotated in dashed box.
    }
    \vspace{-0.15cm}
    \label{fig:intro}

\end{figure}

Despite of these merits, raw images are not widely adopted by common users \wh{due to} large file sizes. 
\wh{In addition, since raw images are unprocessed,}
additional post processing steps, \textit{e.g.}, demosaicing and denoising, are always needed before displaying them.
%
For fast image rendering in practice, a copy of JPEG image is usually saved along with its raw data \cite{dng}.
To improve the storage efficiency, raw-image reconstruction problem attracts more and more attention, \textit{i.e.}, how to minimize the amount of metadata required for de-rendering sRGB images back to raw space.
Classic metadata-based raw image reconstruction methods model the workflow of image signal processing (ISP) pipeline and save \wh{the required parameters in ISP} as metadata \cite{nguyen2016raw}.
\begin{figure}[tbp]
    \centering
    \begin{subfigure}{\linewidth}
    \includegraphics[width=1\linewidth, trim=0 0 0 0, clip]{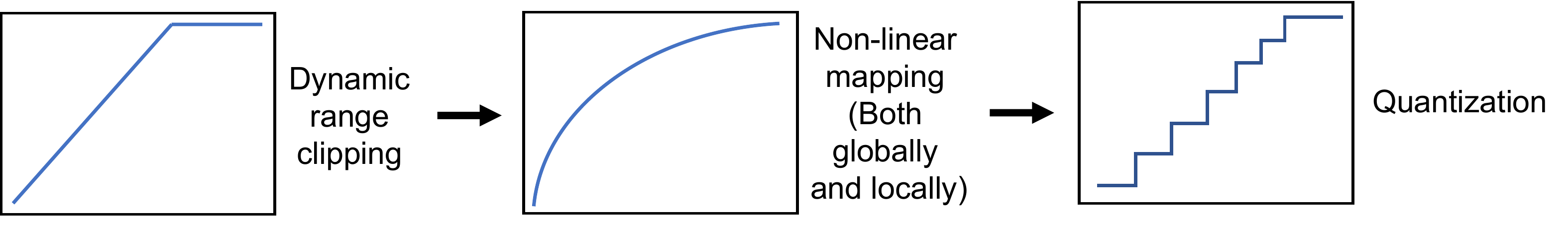}
    \caption{Simplified ISP adopted from \cite{liu2020single}}
    \end{subfigure}\\
    \begin{subfigure}{0.3\linewidth}
    \includegraphics[width=1\linewidth]{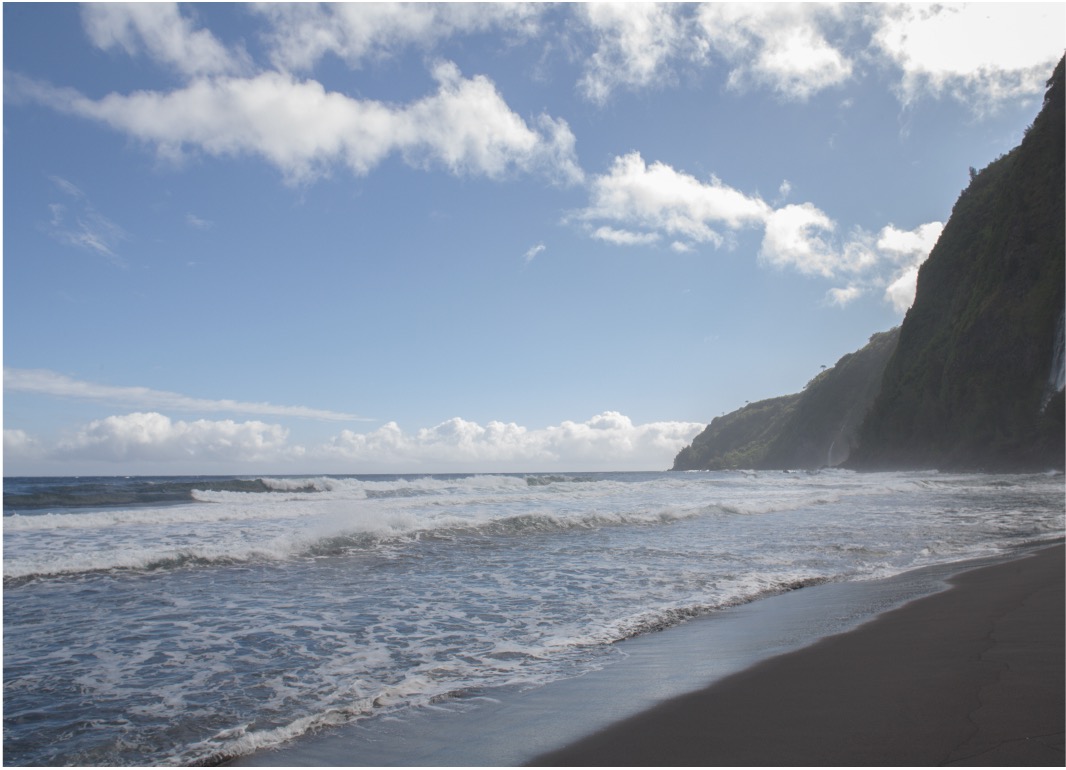}
    \caption{Processed Raw}
    \end{subfigure}
    \begin{subfigure}{0.3\linewidth}
    \includegraphics[width=1\linewidth]{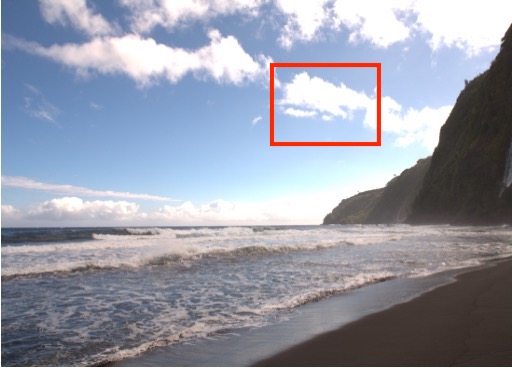}
    \caption{Quantized sRGB}
    \end{subfigure}
    \begin{subfigure}{0.37\linewidth}
    \includegraphics[width=1\linewidth]{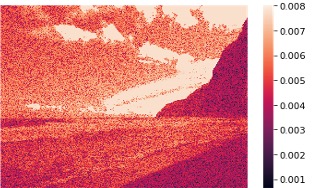}
    \caption{Quantization error map}
    \end{subfigure}
    
    \vspace{-0.1cm}
    \caption{
    An illustration of the information loss caused by the ISP. 
    (a) A simplified ISP \wh{suffers from the} information loss \wh{caused by nonlinear transformations}. 
    (b) Raw image after process to better display the details. 
    (c) Quantized sRGB image after ISP which suffers information loss, \textit{e.g.}, the red bounding box area.
    (d) The quantization error map. 
    As we can see from the above figures, the information loss caused by the quantization is non-uniformly distributed in both over-exposed areas and normally-exposed areas.
    }
    \label{fig:isp}
    \vspace{-0.2cm}
\end{figure}
\wh{To further reduce the storage and computational complexity towards a lightweight and flexible reverse ISP reconstruction, very recent methods focus on sparse sampling of raw image pixels \cite{punnappurath2021spatially, nam2022learning}.}
%
Specifically, \wh{in~\cite{punnappurath2021spatially},} a uniform sampling strategy \wh{is proposed to combine} with \wh{an} interpolation algorithm \wh{that solves} systems of linear equations. 
\wh{The work in~\cite{nam2022learning} proposes a sampling network and approximates the reconstruction process by deep learning to further improve the sampling strategy}.

Though lots of progress has been made, existing sparse sampling based raw image reconstruction methods still face \wh{limitations} in terms of \wh{coding} efficiency and image reconstruction quality. 
%
Specifically, the bit allocation should be adaptive and globally optimized for the image contents, given the non-linear transformation and quantization steps in ISP as shown in Fig. \ref{fig:isp}.
%
For example, the smooth regions of an image can be well reconstructed with much sparser samples, comparing to the texture-rich regions which deserve denser sampling.
%
In constrast, \wh{in existing practices, even for the state-of-the-art method~\cite{nam2022learning} where the sampling is enforced to be locally non-uniform, it is still almost uniform from the global perspective, which causes metadata redundancy and limits the reconstruction performance.}
In addition, very recent works \cite{punnappurath2021spatially, nam2022learning} sample in a fixed sampling space, \textit{i.e.}, raw image space, with a fixed bit depth of sampled pixels, leading to limited representation ability and precision redundancy. 


To address the above issues, instead of adopting a pre-defined sampling strategy or sampling loss, \textit{e.g.}, super-pixel loss~\cite{yang2020superpixel}, we propose a novel end-to-end learned raw image reconstruction framework based on encoded latent features. Specifically, the \textit{latent features} are obtained by minimizing the reconstruction loss and its \wh{bitstream cost} simultaneously.
To further improve the rate-distortion performance, we propose an sRGB-guided context model based on a learnable order prediction network. %
{Different from the commonly used auto-regressive models~\cite{cheng2020learned, minnen2018joint} which encode/decode the latent features pixel-by-pixel in a sequential way, the proposed sRGB-guided context requires \wh{much} fewer steps (reduce by more than $10^6$-fold) with the aid of a learned order mask, which makes the computational cost feasible while maintaining comparable performance.}
Fig.~\ref{fig:intro} compares the proposed raw image reconstruction method with the previous strategies~\cite{cheng2020learned, minnen2018joint}.

\wh{Our contributions are summarized} as follows\wh{,}
\begin{enumerate}
\setlength{\itemsep}{0pt}
\setlength{\parsep}{0pt}
\setlength{\parskip}{0pt}
    \item We propose the first end-to-end deep encoding framework for raw image reconstruction, by fully optimizing the use of stored metadata.
    %
    \item A novel sRGB-guided context model is proposed by introducing two improved entropy estimation strategies, which leads to better reconstruction quality, smaller size of metadata, and faster speed.
    %
    \item \wh{We evaluate our method over popular raw image datasets. The experimental results demonstrate that we can achieve better reconstruction quality with less metadata required comparing with SOTA methods.}
\end{enumerate}

\section{Related Work}
\subsection{Raw image reconstruction}
The current raw image reconstruction works can be categorized into two categories: blind raw reconstruction and raw reconstruction with metadata.

\begin{figure*}[tbp]
    \centering
    \includegraphics[width=\linewidth, trim=0 20 0 10 ,clip]{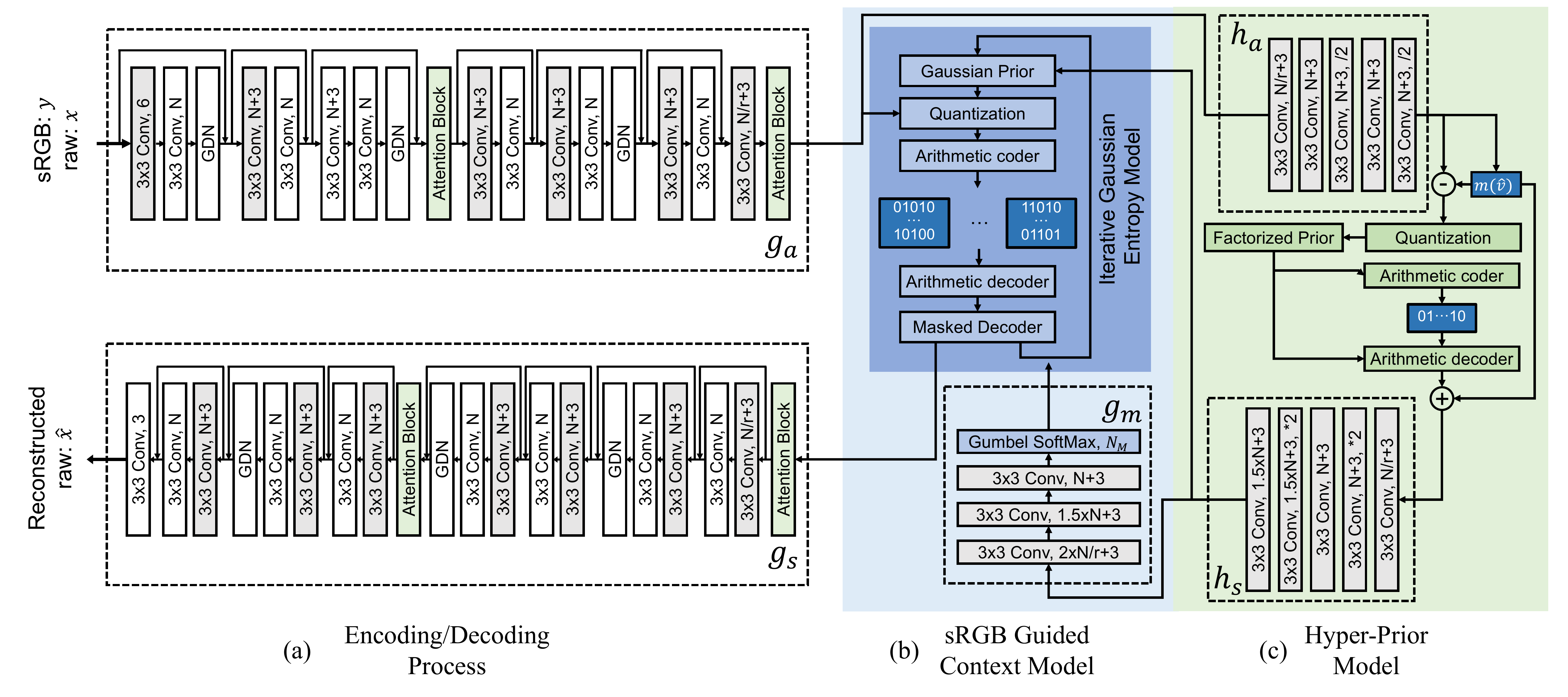}
    \vspace{-0.4cm}
    \caption{
    The overall framework of our method \wh{includes} three parts: 
    (a) The encoding/decoding process \wh{maps} the raw image into the latent space and vice versa. 
    (b) The sRGB-guided context model, which is based on the iterative Gaussian entropy model, \wh{encodes/decodes} the latent variable $\hat{\mathbf{z}}$ into/from \wh{bitstream}. 
    (c) The hyper-prior model \wh{encodes/decodes} the auxiliary variable $\hat{\mathbf{v}}$ into/from the bitstream based on the saved channel-mean value $m(\hat{\mathbf{v}})$.
    The blue blocks with white text represent the saved metadata, and layers in gray represent that the sRGB image is additionally concatenated to their inputs.
    }
    \label{fig:frameowrk}
    \vspace{-0.1cm}
\end{figure*}

\noindent\textbf{Blind raw reconstruction.} Blind raw reconstruction aims to reconstruct the raw image only based on the rendered sRGB image \cite{wang2021deep, zheng2021ultra}. Early works \wh{aim} to recover the linearity of the image by radiometric calibration \cite{debevec2008recovering}. More complex models~\cite{chakrabarti2014modeling, kim2012new, gong2018rank} are subsequently proposed to better describe the workflows of \wh{the} ISP pipeline. With \wh{the} development of the deep-learning, deep-learning based models are rapidly developing. For example, \cite{marnerides2018expandnet} directly learns a mapping from LDR (low dynamic range) to HDR (high dynamic range). \cite{liu2020single} uses three specialized CNNs to reserve the proposed subdivided pipeline from HDR to LDR. Recently, \cite{xing2021invertible} proposes to use an invertible network to learn the mapping between sRGB space and raw space and vice versa. Though great progress \wh{has} been done, due to the information loss during the ISP pipeline, \wh{\textit{e.g.}}, quantization, the fidelity of the reconstructed ones \wh{is} inevitably constrained.

\noindent\textbf{Raw reconstruction with metadata.} To further improve the fidelity of the raw image reconstruction, an alternative way is to save additional metadata to assist the reconstruction \cite{punnappurath2019learning, yuan2011high, nam2022learning}. For instance, \wh{the work in \cite{yuan2011high}} \wh{proposes} to save a low resolution raw file to model the tone mapping curve. \wh{The works in \cite{nguyen2016raw, nguyen2018raw}} propose to save the estimated parameters of the simplified ISP pipeline. \wh{The work in \cite{punnappurath2021spatially}} proposes a spatial aware algorithm that \wh{estimates} the parameters of interpolation during the test time based on saved uniformly sampled raw image pixels. \wh{A recent work~\cite{nam2022learning} improves} the sampling strategy \wh{by sampling} representative raw pixels based on the superpixel. Besides, a UNet is adopted \cite{nam2022learning} to further speed-up the inference process. However, the training of the sampling network is based a pre-defined loss which leads to suboptimal sampling strategy and affects the restoration performance. Different from previous works that usually \wh{save} discrete pixels of raw images, we propose an end-to-end network that can learn to extract necessary metadata in the latent space.

\subsection{Learned image compression}
Recently, a great number of deep learning based image compression methods \cite{balle2016end, li2018learning, hu2021learning} have been proposed and achieve promising results. End-to-end training \wh{is} made possible thanks to the development of differential quantization and rate estimation \cite{theis2017lossy, balle2016end, agustsson2017soft}. Besides, the introduction of \wh{the} contextual model \cite{minnen2018joint, lee2018context} greatly \wh{improves} the compression rate of learned compression models and attracts more and more attention recently. Specifically, \wh{the works} in \cite{minnen2018joint, lee2018context} propose to utilize an autoregressive model to utilize the information that already decompressed from the bitstream. However, due to the nature of the context model, both compress and decompress processes are extremely slow for the image with \wh{high resolution}. To minimize the serial processing, \cite{minnen2020channel} proposes a channel-conditioning and \cite{he2021checkerboard} proposes a checkerboard context model. Besides, though the learning-based image compression exhibits very promising results on the low bpp (bit per pixel) scenarios, the network architecture \wh{needs} to be carefully designed for the settings that require \wh{high fidelity} as shown in \cite{helminger2020lossy, mentzer2020high}.

\section{Methodology}
\subsection{Motivation}
Our goal is to reconstruct the raw image $\mathbf{x}$ which has a linear relationship with the scene radiance based on the sRGB image $\mathbf{y}$ after \wh{the} ISP pipeline.
Due to the operations like quantization and \wh{tone mapping}, the process from the raw image to sRGB image is non-linear and the information loss is spatially non-uniform as shown in Fig. \ref{fig:isp}.
Different from the previous works that uniformly/approximately uniformly save the sparse \textit{raw-pixel} values with a \textit{fixed} number of bits, we propose to learn the coding of information in the \textit{latent space} with an adaptively allocated number of bits for each pixel \wh{in} an end-to-end manner.

As shown in Fig. \ref{fig:frameowrk}, our method aims to \wh{obtain} a compact representation $\mathbf{\hat{z}}$ of the image conditioned on the corresponding sRGB image.
The latent feature $\mathbf{\hat{z}}$ is expected to have necessary information to reconstruct the raw image with high fidelity and \wh{its code-length shall be as small as possible.}
To this end, we propose an sRGB-guided context model which can make better use of decoded information and greatly improve computational efficiency. 
Besides, an improved hyper-prior is proposed to further improve the coding efficiency and reconstruction quality.

\begin{figure*}[t]
    \centering
    \includegraphics[width=\linewidth]{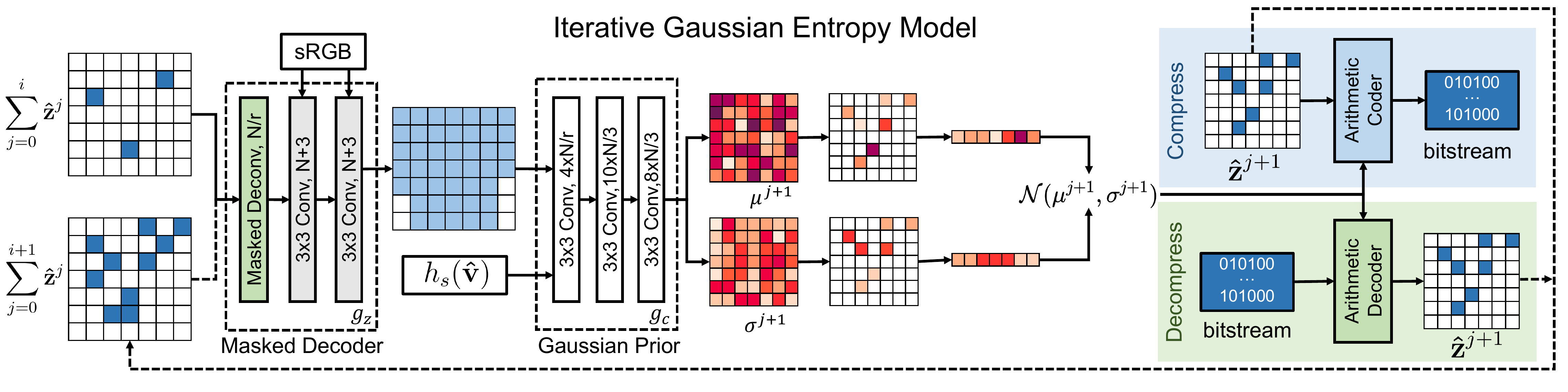}
    \caption{
    An illustration of a step of the proposed iterative Gaussian entropy model. 
    We model the distribution of $\mathbf{\hat{z}}^{i+1}$ based on the existing information $\sum_{j=0}^{i}\mathbf{\hat{z}}^j$ and $h_s(\hat{\mathbf{v}})$ where $\mathbf{\hat{v}}$ is the auxiliary variable from the \wh{hyper-prior}. 
    Then arithmetic coding is used to compress/decompress the latent code $\mathbf{\hat{z}}^{i+1}$ losslessly. The dashed arrows represent the following step.
    }
    \label{fig:context}
    \vspace{-0.0cm}
\end{figure*}

\subsection{The overall of entropy-based coding}
Specifically, our framework can be formulated by
\begin{equation}
\begin{split}
    \mathbf{z} &= g_a(\mathbf{x} , \mathbf{y} ; \phi) \\
    \hat{\mathbf{z}} &= Q(\mathbf{z}) \\
    \hat{\mathbf{x}} &= g_s(\hat{\mathbf{z}},\mathbf{y};\theta), 
\end{split}
\end{equation}
where $\mathbf{z}$ and $\hat{\mathbf{z}}$ are latent codes w/o and w/ quantization. 
$g_a$ and $g_s$ are the analysis and synthesis transforms. $\phi$ and $\theta$ represent the parameters of these two transforms respectively.
$Q$ is the quantization operation. 
We further introduce hypeprior~\cite{balle2018variational} to model the spatial dependencies in $\mathbf{z}$ as follows
\begin{equation}
\begin{split}
\mathbf{v} &= h_a(\mathbf{z}, \mathbf{y}; \phi_h) \\
\hat{\mathbf{v}} &= Q(\mathbf{v}) \\
q_{\hat{\mathbf{z}}|\hat{\mathbf{v}}, \mathbf{y}}(\hat{\mathbf{z}}|\hat{\mathbf{v}}, \mathbf{y}) &\gets h_s(\hat{\mathbf{v}},\mathbf{y};\theta_h),
\end{split} 
\end{equation}
where $h_a$ and $h_s$ represent the auxiliary analysis and synthesis transforms respectively, and $\phi_h$ and $\theta_h$ are the learned parameters of them. The optimization objective that simultaneously minimizes the raw image reconstruction loss and the codelength of latent codes is defined as follows
\begin{equation}
\begin{split}
    \mathcal{L} &= \underbrace{\mathcal{R}(\hat{\mathbf{z}}) + \mathcal{R}(\hat{\mathbf{v}})}_{\text{rate}}  + \lambda \cdot \underbrace{\mathcal{D} (\hat{\mathbf{x}}, \mathbf{x})}_{\text{distortion}} \\
    &= \mathbb{E}[-\log_2{q_{\hat{\mathbf{z}}|\hat{\mathbf{v}},\mathbf{y}}({\hat{\mathbf{z}}|\hat{\mathbf{v}},\mathbf{y}}})] \\ 
    & + \mathbb{E}[-\log_2{q_{\hat{\mathbf{v}}|m(\mathbf{v})}({\hat{\mathbf{v}}|m(\mathbf{v})}})] + \lambda \cdot \mathcal{D} (\hat{\mathbf{x}}, \mathbf{x}), 
\end{split}
\end{equation}
where $m(\mathbf{v})$ represents the mean value of different channels, and $\mathcal{D}$ is the mean square error to measure the reconstruction loss. 
The details of the likelihood estimations of different latent codes will be introduced below.
\subsection{The estimation of the likelihood}
As revealed by the cross entropy $H(p, q)=H(p)+\mathcal{D}_{KL}(p||q)$ that measures the number of extra bits to code the desired distribution $p$ using an estimated one $q$, the key of reducing the code length is to accurately model the distribution of latent codes. 
To this end, we propose to model the distribution of different latent variables with different strategies since they depend on different information.

\begin{figure}[tbp]
    \centering
    \includegraphics[width=0.53\linewidth,trim=0 20 0 0, clip]{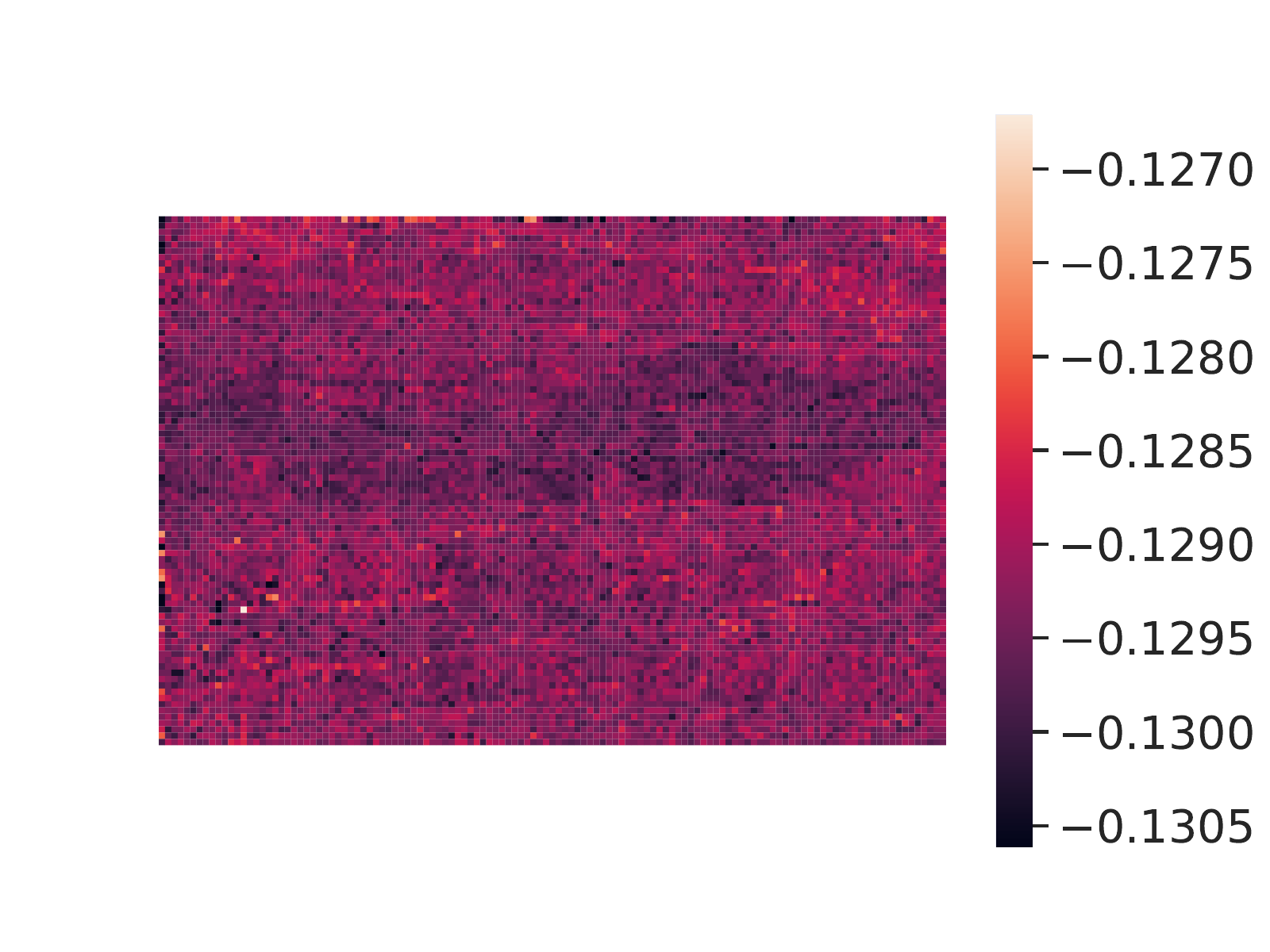}
    \includegraphics[width=0.45\linewidth]{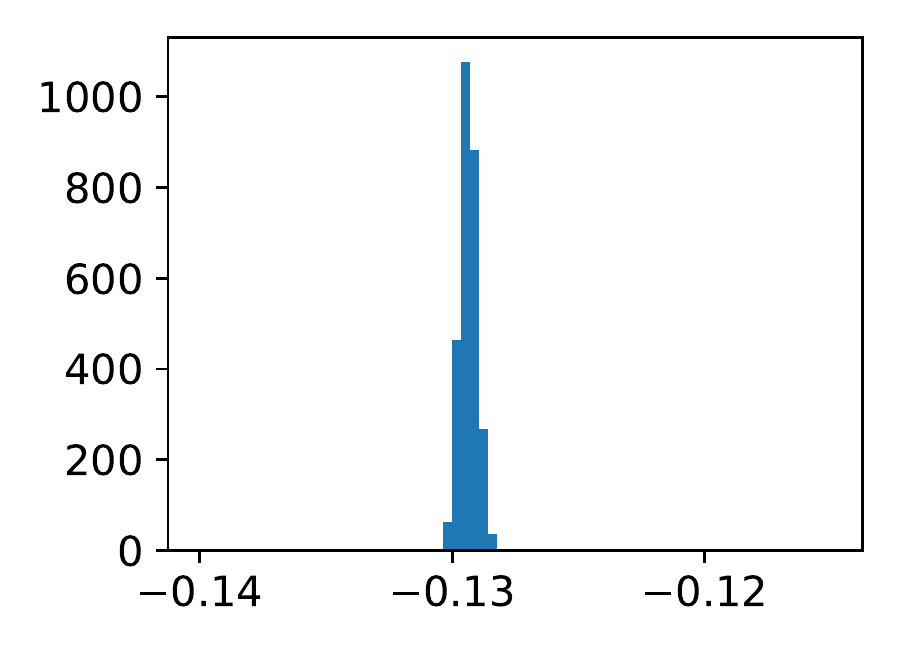}
    \caption{
    \wh{Visualization} of a channel of the auxiliary latent features $\mathbf{v}$ and its value distribution.
    }
    \label{fig:mean_channel}
\end{figure}

Following a similar way of the previous works \cite{balle2018variational, cheng2020learned}, we use a non-parametric, fully factorized density model to encode the auxiliary latent codes $\mathbf{v}$. 
However, due to the limitation of the network design, \wh{\textit{e.g.}}, the domain of the hyper-prior model must be univariate and the network must be monotonic \wh{increasing} \cite{balle2018variational}, we find that there is still lots of redundant information in the auxiliary latent codes $v$ as shown in \wh{Fig. \ref{fig:mean_channel}}. 
To further improve the coding efficiency, we propose to additionally save the mean value of each channel $m(\mathbf{v})$ to the metadata to reduce the redundancy. 
Specifically, we model the conditional distribution $q_{\hat{\mathbf{v}}|m(\mathbf{v})}$ as follows
\begin{equation}
\begin{split}
    &q_{\hat{\mathbf{v}}|m(\mathbf{v})}(\hat{\mathbf{v}}|m(\mathbf{v})) =
    \\ & \prod_{i} [(q_{v_i|m(\mathbf{v})}(\psi^{(i)})*\mathcal{U}(-\frac{1}{2},\frac{1}{2}))(\hat{\mathbf{v}}-m(\mathbf{v}))_i],     
\end{split}
\end{equation}
where $\psi^i$ is the parameters of each univariate conditional distribution $p_{v_i|m(\mathbf{v})}$, and $i$ is the position index.


For the encoding of $\mathbf{\hat{z}}$, previous works show great improvement by introducing the context model, \textit{i.e.}, the already decompressed pixels can help to predict the pixels which are not decompressed yet to further improve the coding efficiency. However, due to its serialization property, the autoregressive model incurs a significant computational cost which is unacceptable in the raw image reconstruction since its high-resolution, \textit{e.g.}, 4000$\times$6000. To improve the computational efficiency while preserve the advantages of the autoregressive model, we propose a novel sRGB-guided context model. More specifically, our proposed context model includes two parts: a learnable order prediction network $g_{m}$ as shown in Fig. \ref{fig:frameowrk}, and an iterative Gaussian entropy model as shown in Fig. \ref{fig:context}.

\noindent\textbf{The learnable order prediction network.}
As the prerequisite of our proposed context model, the order masks of compression/decompression play a significant role. 
To make sampling order masks learnable, we propose a training strategy that \wh{makes} end-to-end training of the whole framework feasible.
Specifically, we utilize \wh{Gumbel}-softmax \cite{jang2016categorical} to make the binary mask derivable for training as follows
\begin{equation}
    M_{i,j}^k = \frac{\exp((\log(m_{i,j}^k)+g^{k}_{ i,j})/\tau)}{\sum_{t=1}^{N}\exp((\log(m_{i,j}^t)+g^{t}_{i,j})/\tau)},
\end{equation}
where $k$ is the index of sampling masks, $N$ is the number of iterations, $\mathbf{g}\in\mathbb{R}^{N\times h\times w}$ is a random matrix i.i.d sampled from Gumbel(0,1) distribution, $\tau$ is a temperature hyper-parameter, and $\mathbf{m}\in\mathbb{R}^{N\times h\times w}$ denotes unnormalized log probabilities predicted by a subnetwork. 
For inference, to make sure that we have exactly the same random vector $\mathbf{g}$ during compress/decompress processes, we add a registered buffer to the model to save a pre-sampled $\mathbf{g}$. 
The pre-sampled $\mathbf{g}$ is then cropped to the same size \wh{as} the $\mathbf{m}$ to generate a set of sparsely sampled $\mathbf{M}^k$.
\begin{figure}[tbp]
    \centering
    \includegraphics[width=0.6\linewidth]{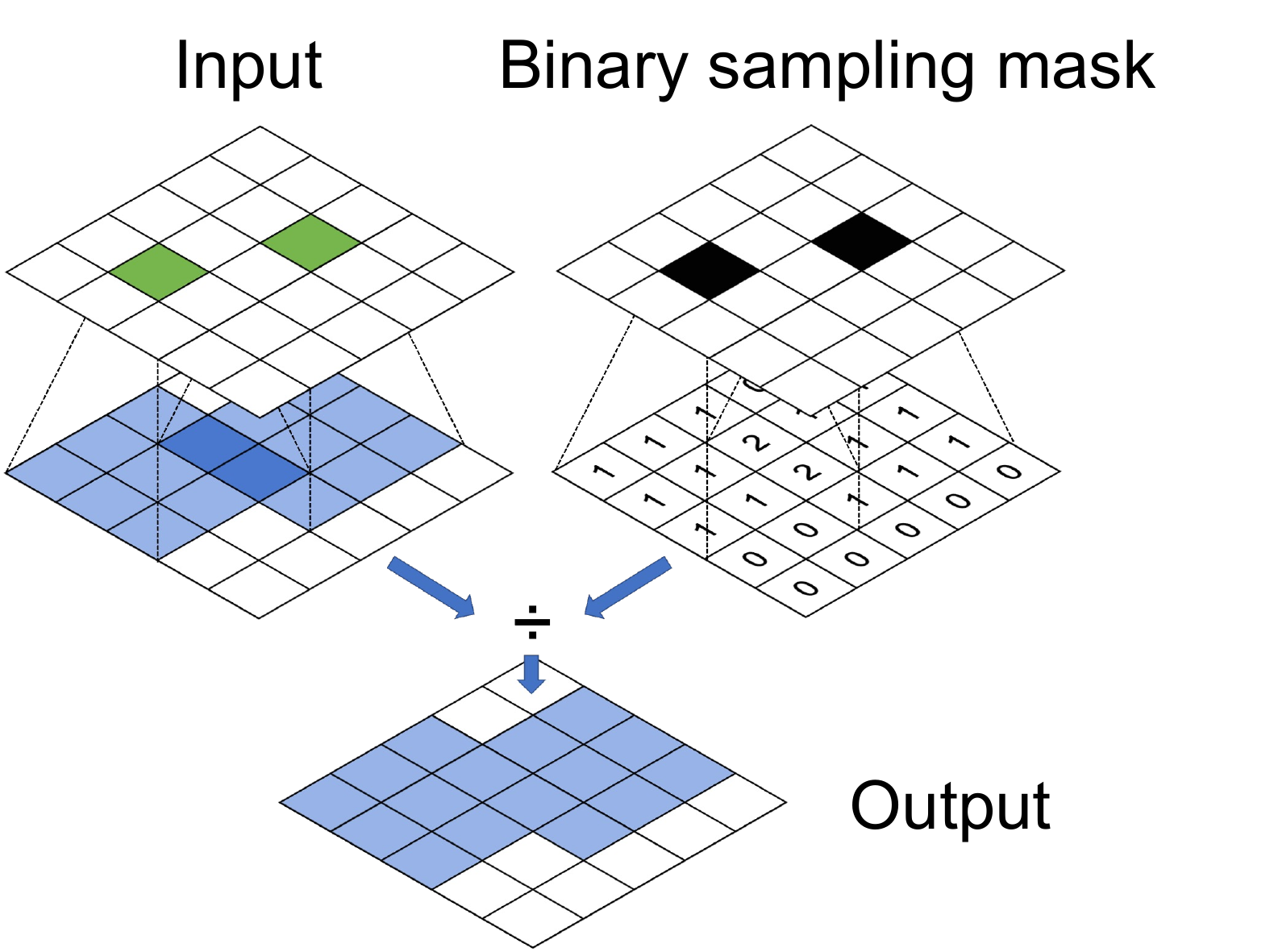}
    \caption{The proposed masked deconvolution layer.}
    \label{fig:mask_deconv}
    \vspace{-0.2cm}
\end{figure}

In addition, we find that \wh{the vanilla convolutional layer cannot} well utilize the information from the randomly sparsely sampled features (can refer to the sampling mask in Fig. \ref{fig:context_vis}).
Therefore, we further propose a new masked deconvolution layer that can alleviate negative impacts from the randomness and sparsity as shown in Fig. \ref{fig:mask_deconv}. 
For an input feature $\mathbf{\hat{z}}$ and its corresponding mask $\mathbf{M}^c=\sum_{i=0}^{k} \mathbf{M}^k$ which records the positions of all already decoded ones, the output $\mathbf{\hat{z}}'$ is as follows
\begin{equation}
    \mathbf{\hat{z}}' = \frac{\text{Deconv}(\mathbf{z})}{\max(1,\text{Deconv}_{ \mathbf{1}}(\mathbf{M}^c))},
\end{equation}
where $\text{Deconv}$ and $\text{Conv}_{ \mathbf{1}}$ are deconvolution layers with stride of 1. Besides,  $\text{Deconv}_{\mathbf{1}}$ is a fixed layer that the weights are all one and the bias is zero. 

\noindent\textbf{The iterative Gaussian entropy model.}
After obtaining the predicted order mask, we can iteratively compress/decompress the information as shown in Fig \ref{fig:context}. Specifically, we use the information from the auxiliary latent variable $\mathbf{\hat{v}}$ and already encoded/decoded partial of $\mathbf{\hat{z}}$ to predict the distribution of the \wh{to-be-processed} part of $\mathbf{\hat{z}}$ as follows
\begin{equation}
    \bm{\mu}^{i+1}, \bm{\sigma}^{i+1} = g_c(g_z((\sum_{k=0}^{i} \mathbf{M}^i) \odot \hat{\mathbf{z}}, y), h_{s}(\hat{\mathbf{v}})),
\label{eq:predict_distri}
\end{equation}
where $\odot$ is a pixel-wise multiplication, $g_z$ is the masked decoder, and $g_c$ is the Gaussian prior module to predict the distribution of $\hat{\mathbf{z}}^{i+1}$ that are not encoded. Then, the likelihood of $\mathbf{\hat{z}}$ is formulated as
\begin{equation}
\begin{split}
    q_{\hat{\mathbf{z}} |\hat{\mathbf{v}},\mathbf{y}}&(\hat{z}_i|\hat{\mathbf{v}},\mathbf{y}) = (\mathcal{N}(\mu^{k_i}_i, \sigma^{k_i}_i) * \mathcal{U}(-\frac{1}{2}, \frac{1}{2}))(\hat{z}_i) \\
    &= c_{\mu^{k_i}_i, \sigma^{k_i}_i}(\hat{z}_i+0.5)  - c_{\mu^{k_i}_i, \sigma^{k_i}_i}(\hat{z}_i-0.5),
\end{split}
\end{equation}
where the subscript $i$ is the index of the pixel position, $k_i$ is the index of \wh{parameter} groups defined in Eq. \ref{eq:predict_distri}, and $c_{\mu^{k_i}_i, \sigma^{k_i}_i}(\cdot)$ is its corresponding cumulative function. 
\begin{figure*}[htbp]
    \centering
    \hspace{-0.4cm}
    \scalebox{0.86}{
    \subcaptionbox{Input\\(8 bit sRGB image)}{
    \includegraphics[height=0.5\linewidth, clip]{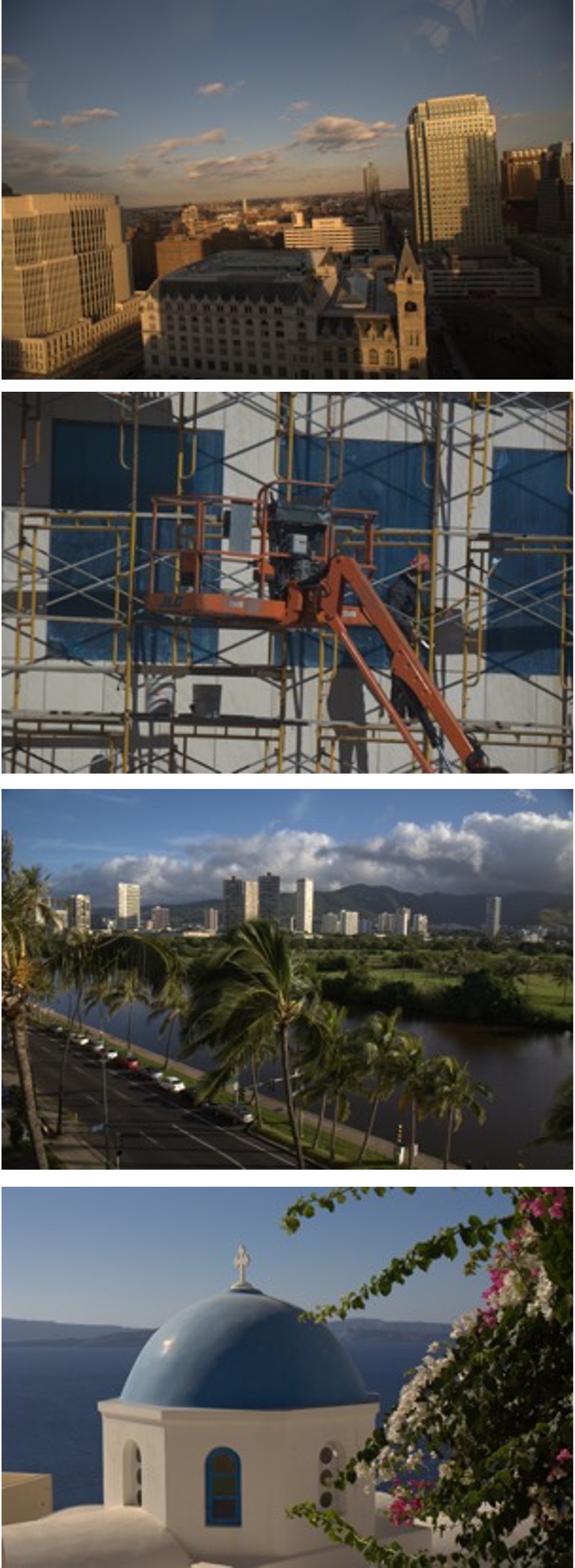}
    }
    \hspace{-0.22cm}
    \subcaptionbox{InvISP~\cite{xing2021invertible}\\(bpp: N/A)}{
    \includegraphics[height=0.5\linewidth, trim=30 0 0 0,clip ]{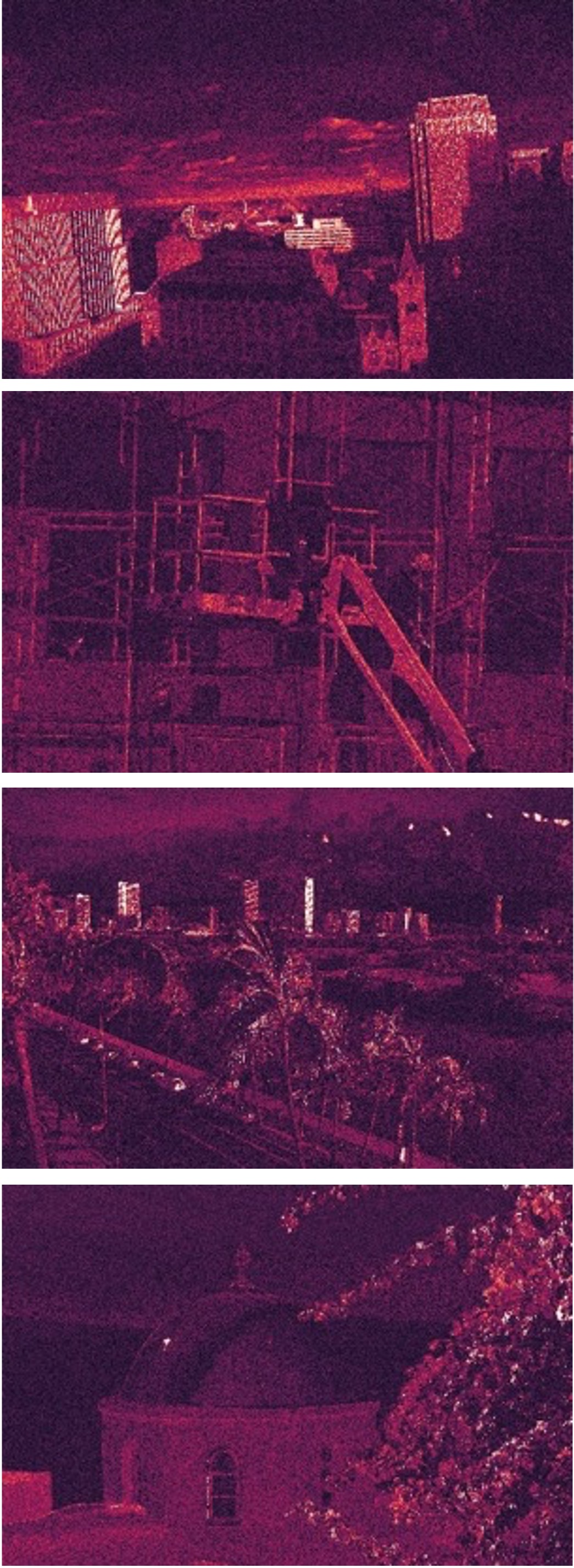}}
    \hspace{-0.2cm}
    \subcaptionbox{SAM~\cite{punnappurath2021spatially} \\(bpp: 9.52e-3)}{
    \includegraphics[height=0.5\linewidth]{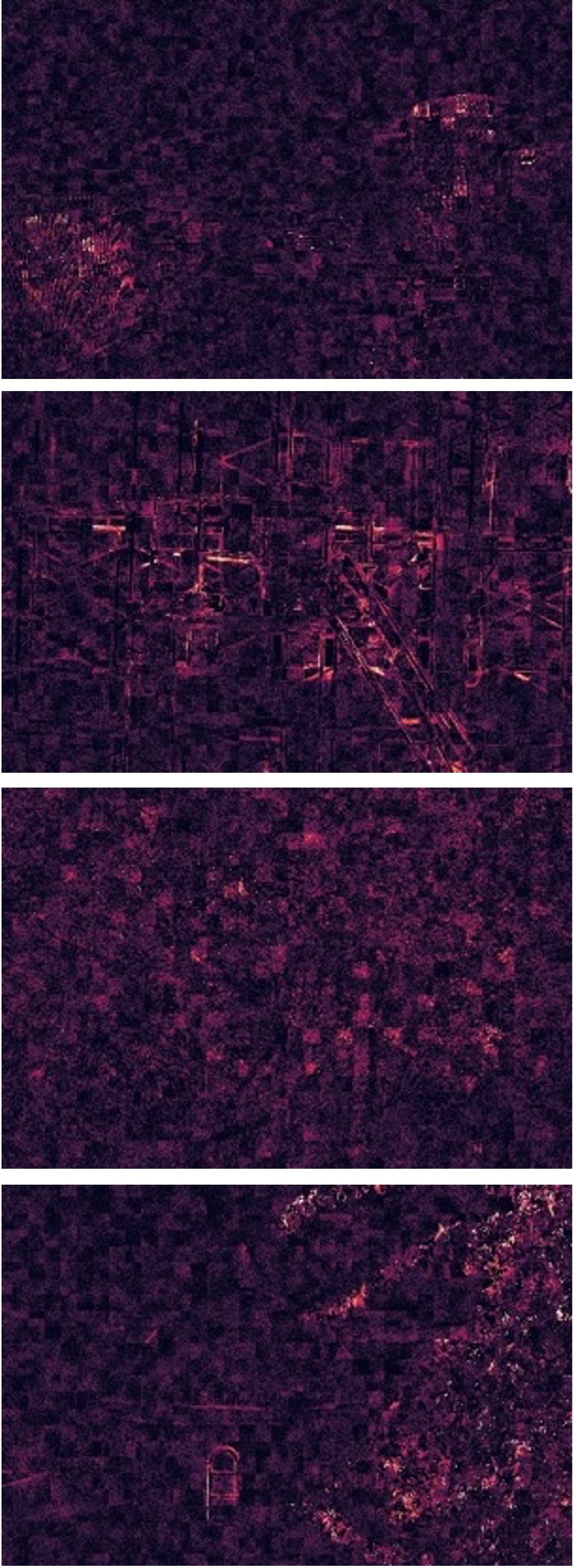}}
    \hspace{-0.2cm}
    \subcaptionbox{Nam~\textit{et al.}\cite{nam2022learning} \\(bpp: 8.44e-1)}{
    \includegraphics[height=0.5\linewidth]{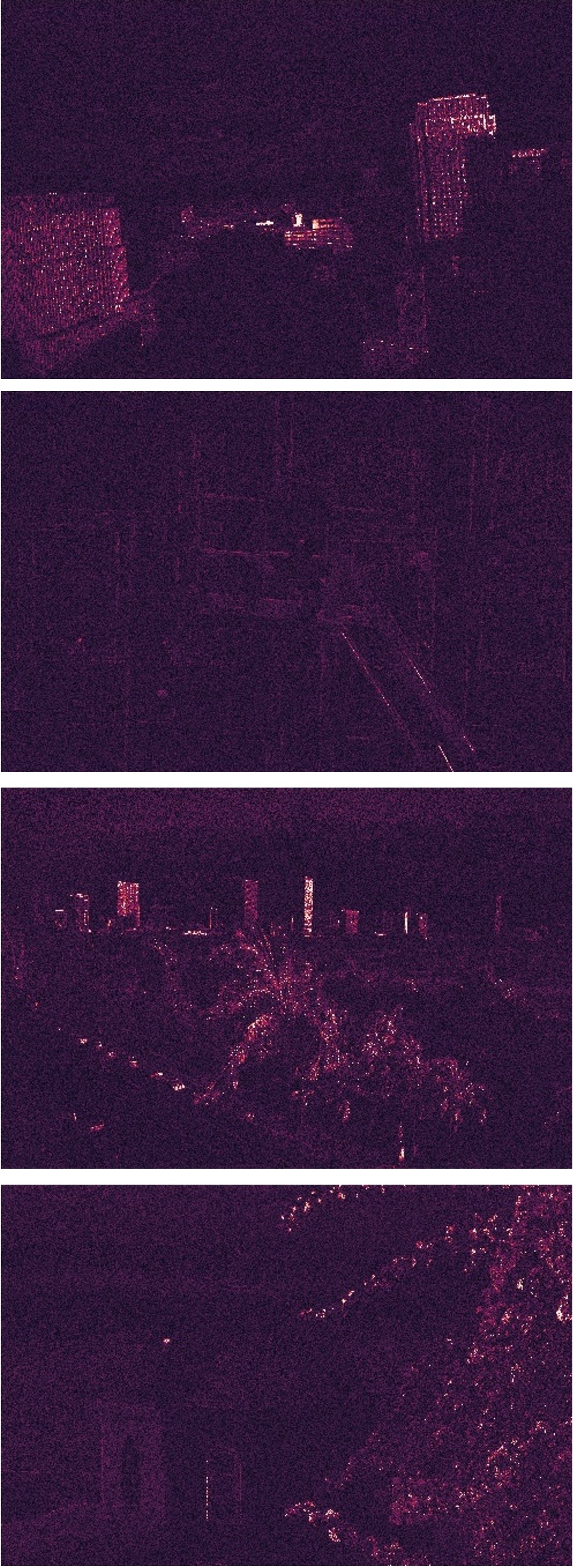}}
    \hspace{-0.2cm}
    \subcaptionbox{Ours\\(bpp: 4.90e-4)}{
    \includegraphics[height=0.5\linewidth]{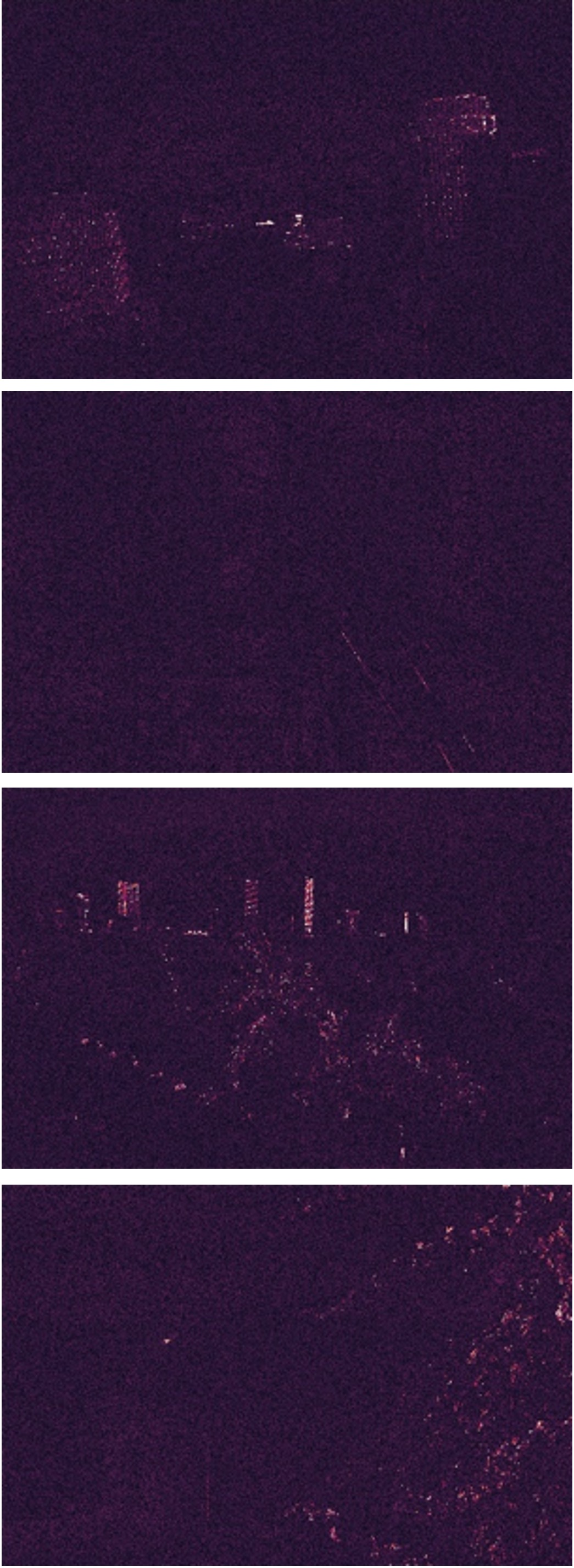}}\hspace{0.cm}
    \includegraphics[height=0.5\linewidth]{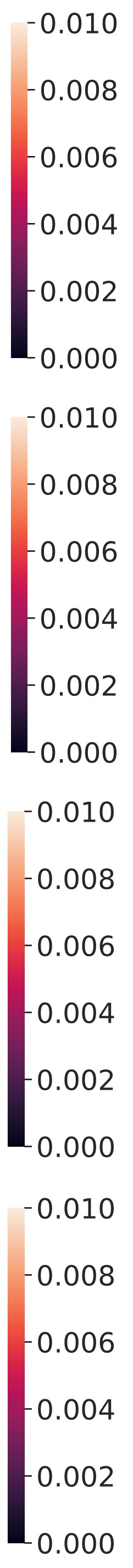}\hspace{-0.1cm}
    \subcaptionbox{Raw image\\(After gamma correction)}{
    \includegraphics[height=0.5\linewidth]{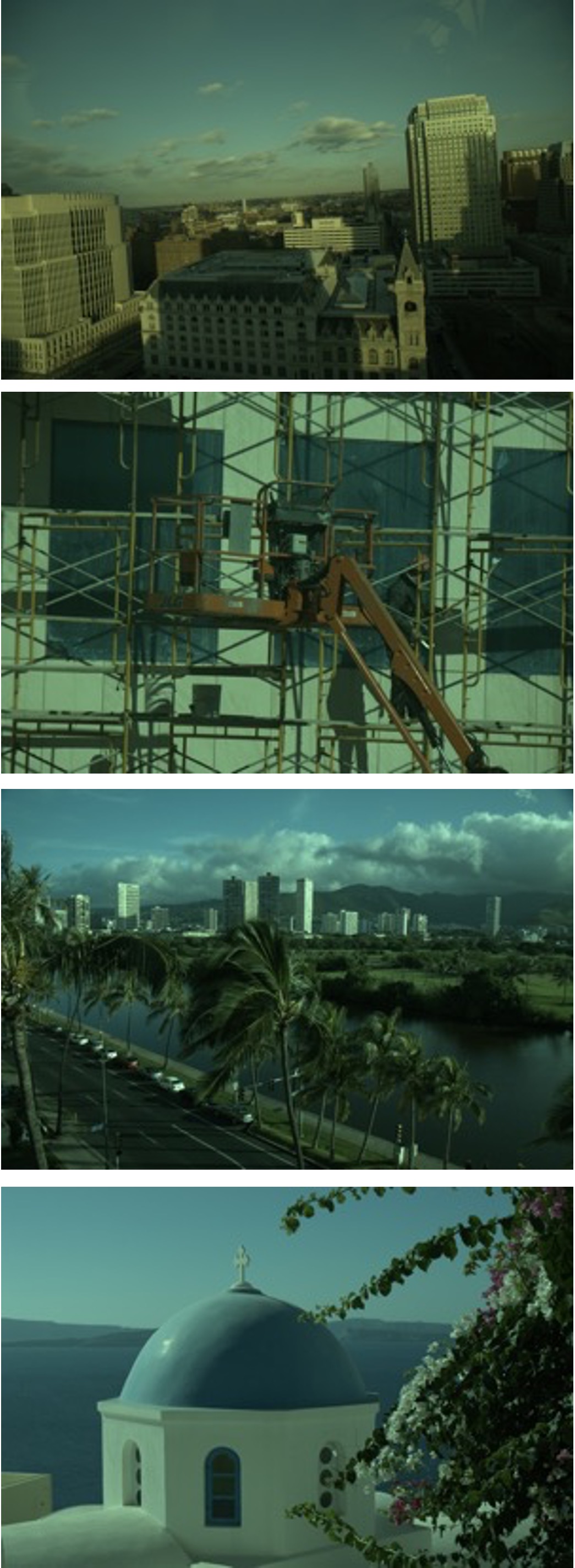}}\hspace{-0.2cm
    }}

    \caption{
    Qualitative comparison of the raw image reconstruction results. We visualize the maximum value of the error among three channels on the pixel level. For better visualization, we apply gamma correction to the raw image to increase the visibility. 
    }
    \label{fig:adobe}
    \vspace{-0.3cm}
\end{figure*}

\section{Experiments}
\subsection{Experimental settings}
\noindent \textbf{Datasets.} 
We utilize two widely-used datasets, NUS dataset \cite{cheng2014illuminant} and AdobeFiveK dataset~\cite{fivek}, to evaluate the effectiveness of our proposed methods. These datasets are all natural images collected from different scenarios and devices. Following previous work \cite{nam2022learning, punnappurath2021spatially}, we use the raw image after demosaic and render sRGB images using a software ISP. 
Specifically, AdobeFiveK dataset~\cite{fivek} includes 5000 photographs taken by different photographers and devices so that it covers a wide range of scenes and lighting conditions. We randomly split the whole dataset into training set and validation set which include 4900 and 100 images respectively.
For the NUS dataset \cite{cheng2014illuminant}, we select the same subsets of devices with the previous work~\cite{nam2022learning}.
%

\noindent \textbf{Baselines.} We compared the proposed method with several SOTA methods, including InvISP ~\cite{xing2021invertible}, SAM~\cite{punnappurath2021spatially}, and Nam \wh{\textit{et al.}} ~\cite{nam2022learning}. Specifically, InvISP~\cite{xing2021invertible} is a SOTA raw image reconstruction model that \wh{utilizes} a single invertible network to learn the mapping from sRGB image to raw image and vice verse. SAM~\cite{punnappurath2021spatially} is a test-time adaptation model that \wh{saves} the uniformly sampled raw pixels as metadata. Nam \wh{\textit{et al.}} ~\cite{nam2022learning} \wh{learn} the sampling process and reconstruction process using two separate neural networks \cite{yang2020superpixel}.

\noindent \textbf{Implementation details.} 
All the code will be released after acceptance. For training, we use a batch size of $1$ and patch size of $1024$ to reduce the I/O time.
Adam is used as the optimizer with a learning rate of 1e-4. 
We train the models for 100 epochs for AdobeFiveK dataset and 200 epochs for NUS dataset. We reduce the learning rate by a factor of 0.1 if there is no improvement in terms of the loss after every 20 epochs. For the sRGB-guided context model, we set $N=2$ for the model conditioned on uncompressed sRGB images and $N=4$ for the compressed JPEG data. 
%
\begin{table}[tbp]
    \centering
    \scalebox{0.93}{
    \begin{tabular}{cccc}
    \toprule
    Method & bpp & PSNR & SSIM \\
    \midrule
    InvISP ~\cite{xing2021invertible} & N/A & 52.69 & 0.99938  \\
    SAM~\cite{punnappurath2021spatially} & 9.566e-4 & 49.61 & 0.99874 \\
    SAM~\cite{punnappurath2021spatially} & 9.5219-3 & 54.76 & 0.99945\\
    Nam \textit{et al.} ~\cite{nam2022learning} & 8.438e-1  & 56.72 & 0.99958 \\
    \rowcolor[HTML]{EFEFEF} Ours (w/o metadata) & N/A &53.03 & 0.99926 \\
    \rowcolor[HTML]{EFEFEF} Ours & \textbf{4.901e-4} & \textbf{58.14} & \textbf{0.99969}  \\
    \bottomrule
    \end{tabular}
    }
    \caption{Quantitative evaluation on AdobeFiveK dataset.}
    \label{tab:adobe}
    \vspace{-0.1cm}
\end{table}

\subsection{Experimental results}
For the evaluation metrics, we utilize PSNR and SSIM \cite{wang2004image} which are widely used to evaluate the reconstruction quality with \wh{the} reference image. 
We also utilize bpp (bit per pixel) to \wh{evaluate} the coding efficiency of the model.

\subsubsection{Results on uncompressed sRGB data.}
\noindent\textbf{Results on AdobeFiveK dataset.}
We report the quantitative evaluation results in Table \ref{tab:adobe}.
As we can see in the table, raw image reconstruction models with metadata can achieve better performance than SOTA raw image reconstruction model without metadata~\cite{xing2021invertible}. Besides, compared with previous metadata-based SOTA methods \cite{punnappurath2021spatially, nam2022learning}, our method achieves better reconstruction quality with lower storage overhead.
Besides, to exclude the effect of network structure, we retrain and evaluate the performance of the model without the help of metadata using the same architecture as ours. Specifically, we set the original input of the raw image to zero, and remove the quantization step and code length loss. 
We find that its reconstruction quality is much lower than the results obtained from the same network with meta-data, which demonstrates the effectiveness of the saved metadata.
Visual comparisons can be found in Fig. \ref{fig:adobe}.

\begin{table*}[t]
\centering
\scalebox{0.96}{
\begin{threeparttable}
\begin{tabular}{cccccccc}

\toprule
\multirow{2}{*}{Method} & \multirow{2}{*}{bpp $\downarrow$} & \multicolumn{2}{c}{Samsung NX2000} & \multicolumn{2}{c}{Olympus E-PL6} & \multicolumn{2}{c}{Sony SLT-A57} \\ \cline{3-8} 
                        &                      & \multicolumn{1}{c}{PSNR $\uparrow$}   & SSIM $\uparrow$  & \multicolumn{1}{c}{PSNR $\uparrow$}  & SSIM $\uparrow$  & \multicolumn{1}{c}{PSNR $\uparrow$}  & SSIM $\uparrow$ \\
\hline
RIR~\cite{nguyen2016raw} & 3.253e-2 & 45.66 & 0.9939 & 48.42 & 0.9924 & 51.26 & 0.9982 \\
SAM~\cite{punnappurath2021spatially} & 7.500e-1 & 47.03 & 0.9962 & 49.35 & 0.9978 & 50.44 & 0.9982\\
Nam \textit{et al.} ~\cite{nam2022learning}~\tnote{1} & 8.438e-1 & 48.08 & 0.9968 & 50.71 & 0.9975 & 50.49 & 0.9973\\
\cite{nam2022learning} w/ fine-tuning  & 8.438e-1 & 49.57 & 0.9975 & 51.54 & 0.9980 & 53.11 & 0.9985\\
\rowcolor[HTML]{EFEFEF}
\rowcolor[HTML]{EFEFEF}
Ours & \textbf{2.887e-4} & \textbf{57.84}\small{$\pm$0.89} & \textbf{0.9997}\small{$\pm$0.00} & \textbf{59.08}\small{$\pm$0.95} & \textbf{0.9998}\small{$\pm$0.00} & \textbf{58.76}\small{$\pm$0.95} & \textbf{0.9997}\small{$\pm$0.00} \\
\bottomrule
\end{tabular}
\begin{tablenotes}
    \item[1] The reason that the bpp of SAM~\cite{punnappurath2021spatially} and Nam \textit{et al.} ~\cite{nam2022learning} is slightly different is that Nam \textit{et al.} ~\cite{nam2022learning} need extra bits to save the locations of sampled raw pixels as shown in Fig. \ref{fig:sampling}.
    \item[2] For the experiments in the gray area, we use a \wh{five-fold} cross validation, \yf{and report the mean and std in the table}.
\end{tablenotes}
\end{threeparttable}
}
\caption{The quantitative results of NUS dataset~\cite{cheng2014illuminant} conditioned on sRGB images. }
\label{tab:nus}
\end{table*}

\noindent\textbf{Results on NUS dataset.}
We also evaluate the performance of models on NUS dataset following a similar evaluation paradigm with Nam \wh{\textit{et al.}} ~\cite{nam2022learning}. 
The results are reported in Table \ref{tab:nus}.
%
As can be seen in the table, we achieve huge performance improvement compared with SOTA method Nam \textit{et al.} ~\cite{nam2022learning} (even compared with the test-time optimization version). 
In addition, the number of bits we need to save as metadata is less than $0.1\%$  of \cite{nam2022learning}.

\subsubsection{Results on compressed sRGB data.}
We further consider a more challenging and realistic setting that we reconstruct the raw image based on compressed JPEG image. 
To evaluate the robustness of our method, we train a single model across different devices and JPEG quality factors. 
The results are reported in Table \ref{tab:nus-compressed}. 
As we can see, our method can adaptively allocate different bpp to JPEG images with different quality factors, \textit{i.e.}, assigning higher bpp to the image with worse JPEG quality. 
Our method achieves the best reconstruction quality with the least metadata compared with previous SOTA methods. 

\begin{table*}[t]
\centering
\setlength\tabcolsep{10pt}
\scalebox{0.925}{
\begin{threeparttable}
\begin{tabular}{ccccccccc}

\toprule
\multirow{2}{*}{Quality} & \multirow{2}{*}{Method} & \multirow{2}{*}{bpp $\downarrow$} & \multicolumn{2}{c}{Samsung NX2000} & \multicolumn{2}{c}{Olympus E-PL6} & \multicolumn{2}{c}{Sony SLT-A57} \\ \cline{4-9} 
& &                      & \multicolumn{1}{c}{PSNR $\uparrow$}   & SSIM $\uparrow$  & \multicolumn{1}{c}{PSNR $\uparrow$}  & SSIM $\uparrow$  & \multicolumn{1}{c}{PSNR $\uparrow$}  & SSIM $\uparrow$ \\
\hline
\multirow{5}{*}{10} & InvISP & N/A & 26.62 & 0.8836 & 29.12 & 0.8980 & 29.12 & 0.9002  \\
& SAM & 9.556e-4 & 24.42 & 0.8946 & 25.24 & 0.9094 & 25.56 & 0.9110 \\
& SAM & 9.522e-3 & 27.94 & 0.9234 & 28.22 & 0.9376 & 27.83 & 0.9374 \\
& Nam \textit{et al.} & 8.438e-1 & 33.06 & 0.9373 & 34.03 & 0.9477 & 34.29 & 0.9506 \\
\rowcolor[HTML]{EFEFEF}
\cellcolor[HTML]{FFFFFF} & Ours & \textbf{7.736e-4}& \textbf{33.13} & \textbf{0.9386}& \textbf{34.04} & \textbf{0.9482}& \textbf{34.31} & \textbf{0.9515} \\ 
\hline
\multirow{4}{*}{30} & InvISP & N/A & 28.71 & 9.9316 & 31.76 & 0.9421 & 30.89 & 0.9459\\
& SAM & 9.556e-4 & 28.88 & 0.9344 & 30.21 & 0.9465 & 29.65 & 0.9458 \\
& SAM & 9.522e-3 & 34.24 & 0.9553 & 35.87 & 0.9648 & 36.12 & 0.9677 \\ 
& Nam \textit{et al.} & 8.438e-1 & 37.21 & 0.9630 & 38.70 & 0.9723 & 39.06 & 0.9750\\
\rowcolor[HTML]{EFEFEF}
\cellcolor[HTML]{FFFFFF} & Ours & \textbf{3.613e-4}& \textbf{37.40} & \textbf{0.9640}& \textbf{38.81} & \textbf{0.9729}& \textbf{39.18} & \textbf{0.9757} \\ %
\hline
\multirow{5}{*}{50} & InvISP & N/A & 30.02 & 0.9416 & 32.91 & 0.9529 & 32.97 & 0.9579\\
& SAM & 9.556e-4 & 30.78 & 0.9448 & 32.41 & 0.9559 & 32.05 & 0.9567 \\
& SAM & 9.522e-3 & 36.32 & 0.9629 & 37.77 & 0.9705 & 38.24 & 0.9739 \\
& Nam \textit{et al.} & 8.438e-1 & 38.34 & 0.9686 & 40.07 & 0.9767 & 40.04 & 0.9797  \\
\rowcolor[HTML]{EFEFEF}
\cellcolor[HTML]{FFFFFF} & Ours & \textbf{3.368e-4}& \textbf{38.67} & \textbf{0.9699}& \textbf{40.33} & \textbf{0.9776}& \textbf{40.73} & \textbf{0.9806}\\ 
\hline
\multirow{5}{*}{70} & InvISP & N/A & 30.86 & 0.9458 & 32.91 & 0.9553 & 32.97 & 0.9592 \\
& SAM & 9.556e-4 & 32.08 & 0.9529 & 34.14 & 0.9620 & 33.90 & 0.9637 \\
& SAM & 9.522e-3 & 37.42 & 0.9684 & 38.96 & 0.9745 & 39.38 & 0.9780 \\
& Nam \textit{et al.} & 8.438e-1 & 39.13 & 0.9724 & 41.01 & 0.9769 & 41.42 & 0.9825 \\
\rowcolor[HTML]{EFEFEF}
\cellcolor[HTML]{FFFFFF} & Ours & \textbf{3.210e-4}& \textbf{39.59} & \textbf{0.9742}& \textbf{41.36} & \textbf{0.9807}& \textbf{41.75} & \textbf{0.9836}  \\
\hline
\multirow{5}{*}{90} & InvISP & N/A & 31.55 & 0.9476 & 33.74 & 0.9598 & 33.68 & 0.9643  \\
& SAM & 9.556e-4 & 34.37 & 0.9663 & 36.60 & 0.9712 & 36.78 & 0.9747 \\
& SAM & 9.522e-3 & 39.17 & 0.9787 & 40.79 & 0.9812 & 41.20 & 0.9843\\
& Nam \textit{et al.} & 8.438e-1 & 40.32 & 0.9782 & 42.33 & 0.9838 & 42.82 & 0.9864  \\
\rowcolor[HTML]{EFEFEF}
\cellcolor[HTML]{FFFFFF} & Ours  & \textbf{2.944e-4}& \textbf{41.19} & \textbf{0.9821}& \textbf{42.98} & \textbf{0.9856}& \textbf{43.43} & \textbf{0.9882}\\
\bottomrule
\end{tabular}
\end{threeparttable}
}
\caption{The quantitative results of NUS dataset~\cite{cheng2014illuminant} conditioned on the compressed JPEG image with different quality factors. }
\label{tab:nus-compressed}
\vspace{-0.2cm}
\end{table*}

    

\subsection{Ablation study}
\begin{figure}[tbp]
    \centering
    \hspace{-0.4cm}
    \scalebox{0.85}{
    \subcaptionbox{input}{
    \includegraphics[height=0.5\linewidth, clip]{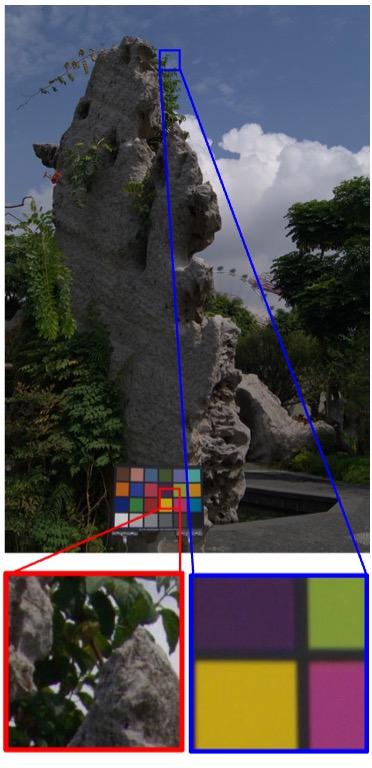}
    }
    \hspace{-0.2cm}
    \subcaptionbox{SAM~\cite{punnappurath2021spatially}\\
    bpp: 9.52e-03\\
    PSNR: 55.68}{
    \includegraphics[height=0.5\linewidth, trim=30 0 0 0,clip ]{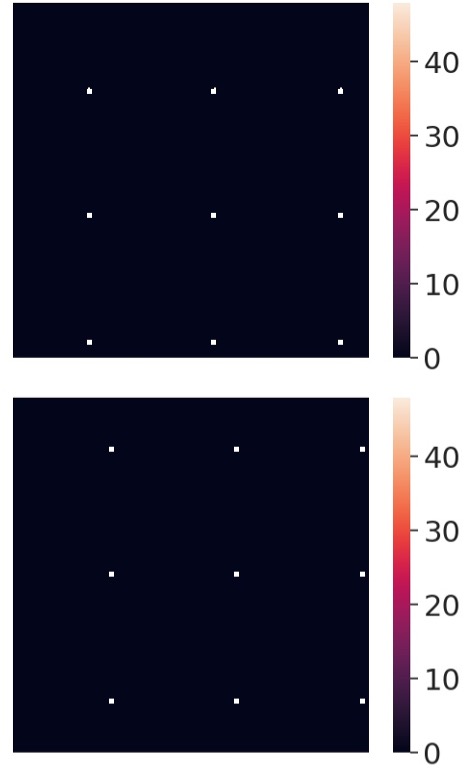}}
    \hspace{-0.2cm}
    \subcaptionbox{Nam \textit{et al.}. \cite{nam2022learning}\\
    bpp: 8.44e-1\\
    PSNR: 57.73}{
    \includegraphics[height=0.5\linewidth]{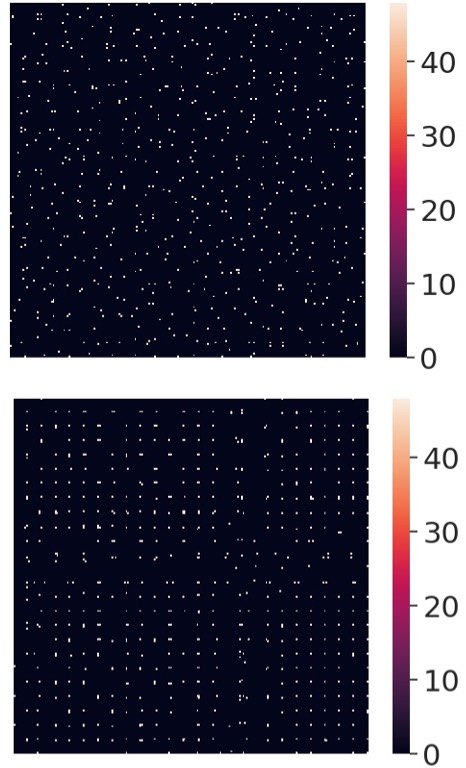}}
    \hspace{-0.2cm}
    \subcaptionbox{Ours\\
    bpp: 2.90e-4\\
    PSNR: 60.04}{
    \includegraphics[height=0.5\linewidth]{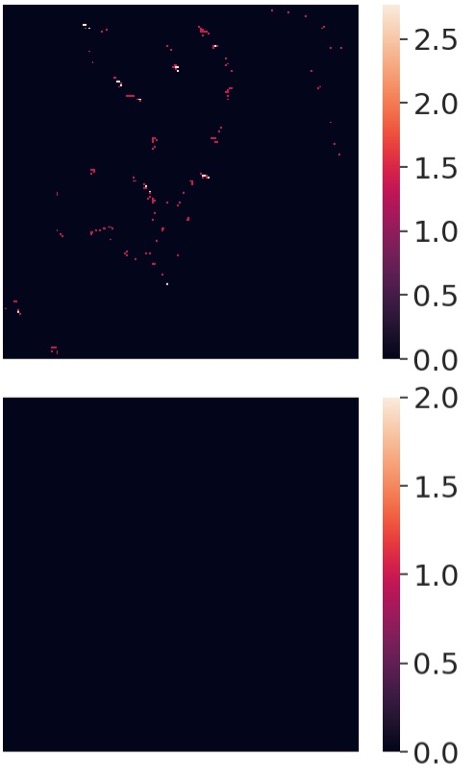}}\hspace{-0.2cm}}
    \vspace{-0.2cm}
    \caption{
    The comparison of the bits allocation. 
    (a) The input 8-bit sRGB image. 
    (b)-(d) The bits allocation maps of the red bounding box area (the first row) and the blue area (the second row). 
    For better visualization, we enlarge the size of sampled pixels in (b). 
    It is worth noting that for (b)-(c), each sampled raw pixel needs 48 bits to save. 
    \textit{Best zoom in for more details.}
    }
    \label{fig:sampling}
\end{figure}

\noindent\textbf{The comparison of bits allocation.}
One of the main advantages of the proposed method is that we could learn the bits allocation in an end-to-end manner. 
As shown in Fig. \ref{fig:isp}, the information loss is non-uniform so a good bits allocation algorithm is the core of metadata-based raw image reconstruction algorithms.
To this end, we visualize the bits allocation of both current SOTA methods and the proposed method in Fig. \ref{fig:sampling}.
In \cite{punnappurath2021spatially}, the metadata are uniformly sampled in  the raw pixel space, which leads to \wh{redundancy}. 
Although Nam \wh{\textit{et al.}} ~\cite{nam2022learning} propose a superpixel based sampling network, the training of reconstruction and sampling networks are \wh{separated into} two phases.
In addition, even if its sampling is locally non-uniform, it still remains uniform globally, which limits the coding efficiency and reconstruction quality. 
As we can see in the figure, our method can adaptively allocate different bits to different areas. 
Specifically, for the flat area, \textit{e.g.} flat area in the blue bounding box, our methods utilize few bits to encode. While for the area with more complicated context, \textit{e.g.} boundary area in the blue bounding box, our method allocates relatively more bits.
Besides, even for areas where we allocate bits, the need of bits is much lower than the methods \wh{that sample} in the raw \wh{pixel} space.

\begin{table}[htbp]
    \centering
    \scalebox{0.82}{
    \begin{tabular}{c|ccc|ccc}
    \toprule
         & \multicolumn{3}{c|}{Ours w/o $m(\mathbf{v})$} & \multicolumn{3}{c}{Ours}\\ \cline{2-7}
         & bpp $\downarrow$ & PSNR & SSIM & bpp $\downarrow$ & PSNR & SSIM \\
    \midrule
      Samsung   & 2.92e-4 & 57.50 & 0.9997 & 2.88e-4 & 57.79 & 0.9997\\
      Olympus & 2.93e-4 & 58.93 & 0.9997 & 2.90e-4 & 59.35 & 0.9997 \\
      Sony & 2.90e-4 & 59.05 & 0.9997 & 2.89e-4 & 59.24 & 0.9997 \\
      \rowcolor[HTML]{EFEFEF} Mean & 2.92e-4 & 58.66 & 0.9996 & \textbf{2.89e-4} & \textbf{58.79} & \textbf{0.9997} \\
    \bottomrule
    \end{tabular}
    }
    \caption{The ablation study on the proposed modeling of $\mathbf{\hat{v}}$. }
    \label{tab:likelihood_estimation}
\end{table}

\noindent\textbf{The hand-crafted metadata of $\mathbf{v}$.}
To verify the effectiveness of our proposed modeling of the hyper-prior variable $\mathbf{\hat{v}}$.
We compare the models trained w/ and w/o the hand-crafted metadata $m(\mathbf{v})$. 
We keep other settings the same as in Table \ref{tab:nus-compressed} and evaluate the models on a fold of NUS dataset. 
The results are reported in Table \ref{tab:likelihood_estimation}.
As we can see in the table, there are improvements in terms of both the bpp and the reconstruction quality \wh{that benefited} from the more accurate value of $\mathbf{v}$ and alleviated redundancy.


\begin{figure}
\includegraphics[width=\linewidth]{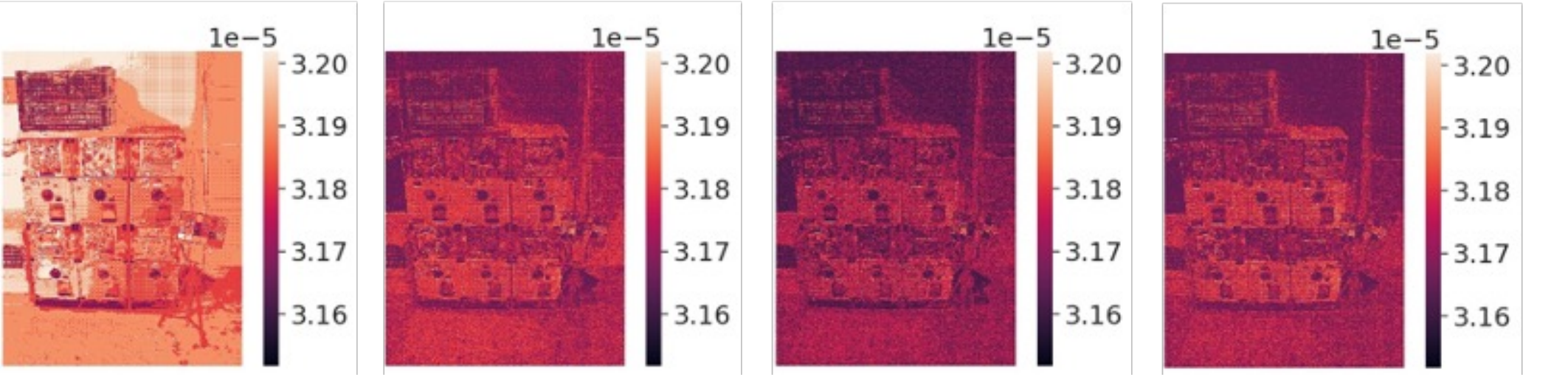}
\caption{
The visualization of each step of the sRGB guided context model 
(a) The sampling masks from $\mathbf{M}^0$ to $\mathbf{M}^3$ respectively. The sampling rate and the bytes of \wh{the} encoded string are displayed on the right side of the mask.
(b) The bpp maps of the latent variable $z$ based on the already decoded information. 
Specifically, the $i_{th}$ (range from $0$ to $3$) bpp map is estimated using the information of $\sum_{k=-1}^{i-1} \mathbf{M}^{k} \odot \mathbf{\hat{z}}$, where $\mathbf{M}^{-1}$ is an all zero mask.
}
\label{fig:context_vis}
\end{figure}

\begin{figure}
    \centering
    \hspace{-0.5cm}
    \scalebox{1.08}{
    \includegraphics[width=0.49\linewidth, trim=0 0 0 0, clip]{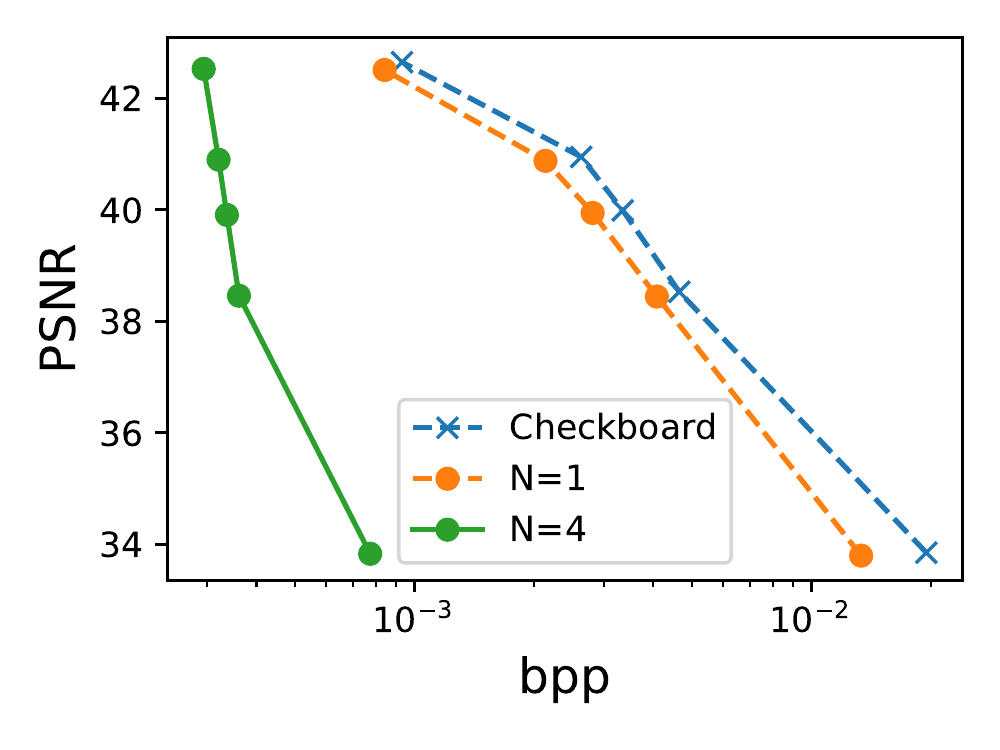}\hspace{-0.22cm}
    \includegraphics[width=0.49\linewidth]{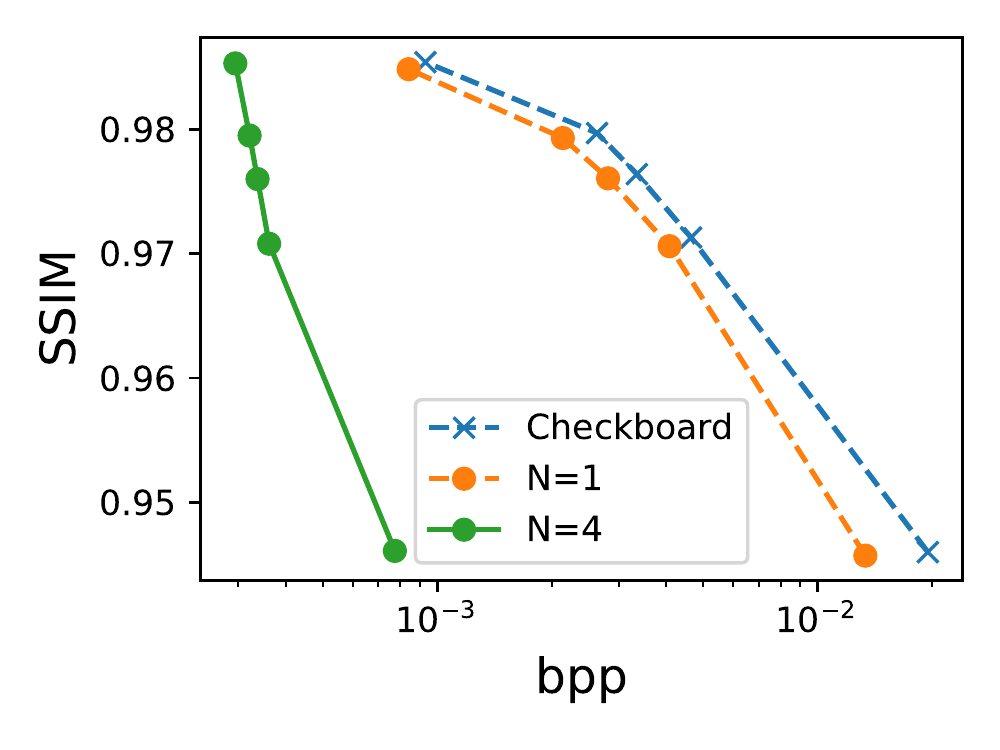}\hspace{-0.22cm}}
    \caption{A comparison of models trained with different steps of the proposed sRGB-guided context model and He \textit{et al}. \cite{he2021checkerboard}. The models are evaluated using JPEG images with varying quality factors (10, 30, 50, 70, 90). For all models, PSNR and SSIM decrease monotonously, with the lowering of the conditioned JPEG quality.}
    \label{fig:rgb-guided}
    \vspace{-0.1cm}
\end{figure}

\noindent\textbf{The sRGB-guided context model.} 
To better understand how the proposed context model works, we visualize each step of encoding/decoding $\mathbf{z}$ in Fig. \ref{fig:context_vis} (a).
As we can see, the order of \wh{the} compress/decompress process is highly-related to the \wh{context} of the image, which demonstrates that our proposed context model can well utilize the information from the sRGB image. 
The sparse sampling mask can help \wh{a} model better predict the distribution of the adjacent pixels that have not been compressed/decompressed. 
In addition, our method gradually increases the number of sampled pixels in the latent space as more and more pixels are available to help to predict the distribution of the unseen ones, which leads to better coding efficiency.
As shown in Fig. \ref{fig:context_vis} (b), we can get more accurate estimation of the likelihood (\wh{\textit{i.e.}}, smaller bpp) of $\mathbf{z}$ with the help of already decoded $\mathbf{\hat{z}}$. 
We also quantitatively evaluate the effectiveness of our proposed sRGB-guided context model as shown in Fig. \ref{fig:rgb-guided}. 
We compare our proposed sRGB-guided context model with He \textit{et al.} \cite{he2021checkerboard} which \wh{proposes} an improved context model to obtain faster speed. 
For a fair comparison, we directly replace our proposed context model (Fig. \ref{fig:frameowrk} (b)) with the checkboard one in He \wh{\textit{et al.}}
\cite{he2021checkerboard} and keep other settings the same. As we can see, all models achieve very similar reconstruction quality and our method achieves much lower bpp.

\section{Conclusion}
In this work, we propose a novel framework for the raw image reconstruction with learned compact metadata. Specifically, the end-to-end learned coding technologies we incorporated can encode the metadata in the latent space with an adaptive bits allocation strategy, which achieves better reconstruction quality and higher coding efficiency. Our further proposed novel sRGB-guided context model leads to
better reconstruction quality, smaller size of metadata,
and faster speed. We evaluate our method on widely-used datasets and the results demonstrate that our method significantly improve performance over prior methods, \textit{i.e., } we achieve better reconstruction quality and smaller size of metadata.
{\small
\bibliographystyle{ieee_fullname}
\bibliography{egbib}
}
\end{document}